\pdfoutput=1
\documentclass[11pt]{article}
\usepackage[final]{acl}
\usepackage{times}
\usepackage{latexsym}
\usepackage[T1]{fontenc}
\usepackage[utf8]{inputenc}
\usepackage{microtype}
\usepackage{inconsolata}
\usepackage{amsmath}
\usepackage{mathtools}
\usepackage{bm}
\usepackage{amssymb}
\usepackage{graphicx}
\usepackage{graphicx,xcolor}
\usepackage{booktabs}
\usepackage {colortbl,array}
\usepackage{multirow}
\usepackage{enumitem}
\usepackage{tabularx}
\usepackage{ragged2e}
\usepackage{xcolor}
\definecolor{mygreen}{RGB}{0,177,97}
\usepackage[most]{tcolorbox}
\newcommand{\TColorBox}[2]{
\begin{tcolorbox}[enhanced, breakable,
  colback=blue!5!white,
  colframe=blue!75!black,
  fonttitle=\bfseries,
  title=#1, 
  boxrule=0.5pt,
  sharp corners,
  left=2mm, right=2mm, top=1mm, bottom=1mm,
  before upper={\let\\\par},
]
#2
\end{tcolorbox}
}

\title{
  \centering
  \makebox[\textwidth][c]{
    \parbox[b]{0.8\textwidth}{\centering
      \textbf{\textsc{Rigel}: \Large Self-Distilled Score Adaptation \\for Image and Video Captioning Evaluation}
    }
  }
}
\author{ \begin{tabular}{@{}c@{\hspace{5mm}}c@{\hspace{5mm}}c@{}} \textbf{Shuitsu Koyama} & \textbf{Kazuki Matsuda} & \textbf{Yuiga Wada} \\ \textbf{Shinnosuke Hirano} & \textbf{Daichi Yashima} & \textbf{Komei Sugiura} \end{tabular} \\ Keio University \\ {\fontsize{10.5pt}{11pt}\selectfont \texttt{ \{koyamashu3, k2matsuda0, yuiga, shinhirano, ydaichi1207, komei.sugiura\}@keio.jp }} }
\begin{document}
\maketitle

\begin{abstract}
Automatic evaluation of image and video captioning is essential for benchmarking multimodal systems, although standard evaluation metrics show limited alignment with human judgments. 
Recent approaches using large language models (LLMs), commonly referred to as LLM-as-a-Judge, have improved alignment with human judgments but still suffer from a mismatch between large-vocabulary language modeling and evaluation over a small label set.
To address this, we propose \textsc{Rigel}, an automatic evaluation metric for image and video captioning, based on self-distilled score adaptation.
The metric employs an evaluation-specific scoring head distilled from a frozen LLM, which captures judgment signals in a task-aligned space without relying on large-vocabulary token sets.
We then refine the LLM backbone with human judgment data. 
To train \textsc{Rigel}, we constructed the Vid-Lepus dataset, which contains $3,338$ video clips, $33,380$ reference captions, and $5,637$ candidate captions.
Experiments on multiple benchmarks show that \textsc{Rigel} outperforms state-of-the-art metrics, achieving over 10-point improvements on ActivityNet-Fact in the reference-free setting. Our project page is available at \url{https://rigel-mnghv.kinsta.page/}
\end{abstract}

\section{Introduction}

Automatic evaluation of image and video captioning is essential for benchmarking multimodal systems~\cite{qwen-vl, internvl, gemini}.
However, standard evaluation metrics show limited alignment with human judgments.
Traditional metrics such as BLEU~\cite{bleu} and METEOR~\cite{meteor} rely on lexical or semantic matching and correlate weakly with human judgments~\cite{clipscore, pac-s, pac-spp}.
Data-driven metrics based on vision language models (e.g., Polos~\cite{polos}) improve correlation but capture surface-level similarity rather than fine-grained dimensions such as descriptiveness, relevance, and fluency.

\begin{figure}[t]
    \centering
    \includegraphics[width=1.0\linewidth]{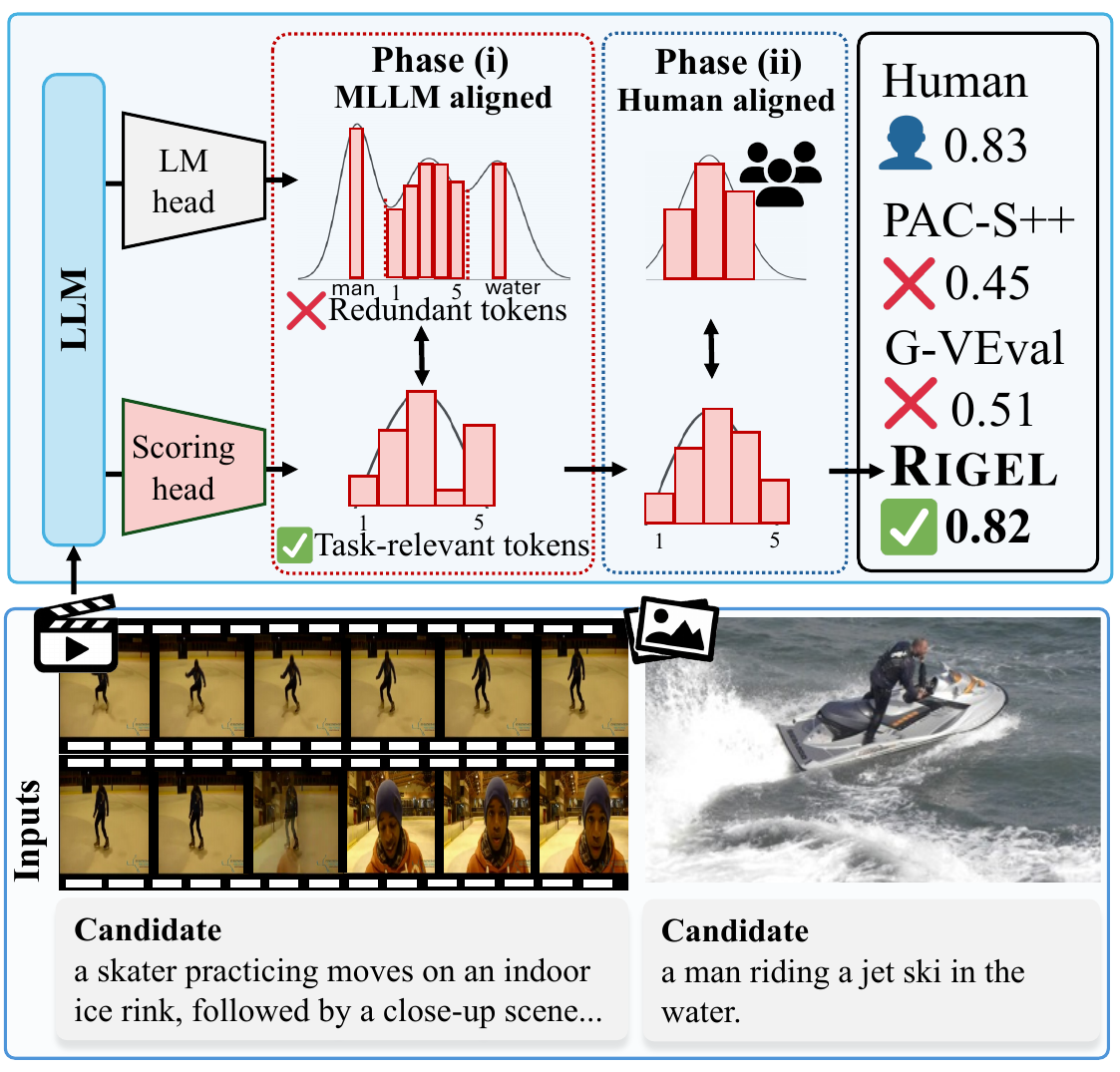}
   \caption{\textbf{Overview of \textsc{Rigel}.} A two-phase framework for human-aligned caption evaluation. In Phase 1, an evaluation-specific scoring head is distilled from a frozen large language model (LLM) to map hidden representations to ordinal judgment scores, alleviating the mismatch between the LM vocabulary and the ordinal label set in the original language modeling (LM) head. In Phase 2, the LLM backbone is refined using human judgments (e.g., scores ranging from 1 to 5) while freezing the scoring head's parameters, yielding task-aligned evaluations.}
    \label{fig:eye-catch}
    \vspace{-3mm}
\end{figure}

\begin{figure}[t]
    \centering
    \includegraphics[width=1\linewidth]{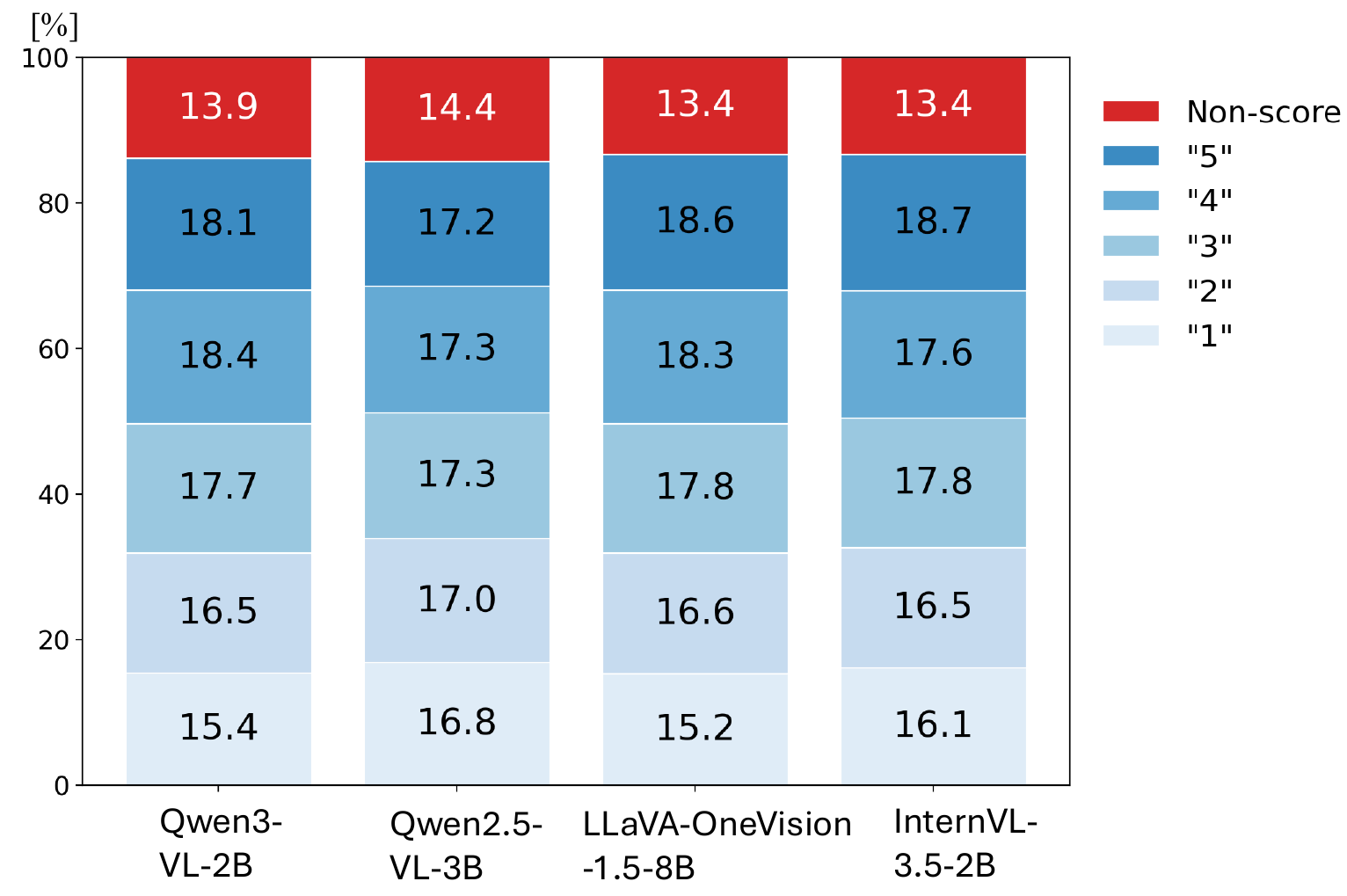}
   \caption{Logit distributions over score tokens (``1''–``5'') and non-score tokens on the Spica dataset~\cite{pearl}. Non-score tokens exhibit logit magnitudes comparable to those of score tokens across the four models Qwen3-VL-2B~\cite{qwen3vl}, Qwen2.5-VL-3B~\cite{qwen25vl}, LLaVA-OneVision-1.5-8B~\cite{llavaov1.5}, and InternVL-3.5-2B~\cite{internvl35}. This observation provides partial evidence for our claim that non-score tokens function as noise in score prediction. The implementation details are provided in Appendix \ref{additional_logit_distribution}.
   }
    \label{fig:token_logits}
    \vspace{-4mm}
\end{figure}

Recent LLM-based approaches such as FLEUR~\cite{lee2024fleur} and G-VEval~\cite{tong2025gveval} yield more interpretable judgments~\cite{survey-llm-judge}.
However, these methods are limited by their reliance on predefined token sets for scoring.
The language modeling (LM) head operates over a large vocabulary set $\mathcal{V}$ ($|\mathcal{V}| \sim 10^5$), while evaluation requires prediction over a small ordinal label set $\mathcal{M}$ (\(|\mathcal{M}| \ll |\mathcal{V}|\)).
The distributions of LM-head logits assigned to score tokens (``1''–``5'') and the remaining vocabulary are presented in Fig.~\ref{fig:token_logits}. We observe that the logits assigned to task-irrelevant tokens are comparable in magnitude to those of the score tokens.
This indicates that the LM head predicts scores while assigning many logits to task-irrelevant tokens, which can introduce noise and degrade evaluation performance.

Based on the above analysis, we propose \textsc{Rigel}, an automatic evaluation metric for image and video captioning, which addresses the misalignment between the LM vocabulary and the ordinal label set through self-distilled score 
adaptation framework (Fig.~\ref{fig:eye-catch}).
The framework consists of the following two phases:
First, the metric employs an evaluation-specific scoring head distilled from a frozen LLM, which captures judgment signals in a task-aligned space without relying on large vocabulary token sets.
Second, we refine the LLM backbone with human judgment data while freezing the head's parameters.
Our empirical results show that each phase contributes independently to the final performance (Section~\ref{sec:ablation}).

To train our proposed metric, training data that pair candidate captions with human judgments are required. However, existing datasets~\cite{factvc, emscore} often lack this form of paired supervision, making it challenging to develop supervised evaluation metrics. To address this limitation, we constructed Vid-Lepus, a dataset for training video captioning evaluation metrics. Vid-Lepus comprises video clips, reference captions, candidate captions, and corresponding human judgments.

The main contributions of this study are ~summarized as follows:
\begin{itemize}[leftmargin=*, itemsep=0pt]
    \item We propose \textsc{Rigel}, a unified evaluation metric for image and video captioning (Section~\ref{sec:Rigel}), addressing the vocabulary-label mismatch between the LM head vocabulary and ordinal label set.
    \item We introduce self-distilled score adaptation framework. Our empirical results show that our metric, \textsc{Rigel}, outperforms existing metrics on standard benchmarks (Section \ref{sec:experiments}).
    \item We introduce Vid-Lepus, a dataset for training human-aligned metrics for video captioning, featuring $3,338$ video clips, $33,380$ reference captions, $5,637$ candidate captions, and $14,802$ human judgments (Section \ref{sec:myvatex}).
\end{itemize}

\section{Related Work}
\label{sec:related_work}
\paragraph{Image captioning metrics.}
Automatic evaluation metrics for image captioning, such as BLEU~\cite{bleu}, METEOR~\cite{meteor}, ROUGE~\cite{rouge}, CIDEr~\cite{cider}, and SPICE~\cite{spice}, along with extensions such as CIDEr-R~\cite{cider-r} and JaSPICE~\cite{jaspice}, have traditionally relied on reference-based lexical or semantic matching.
However, these metrics often correlate only weakly with human judgments, particularly when captions are semantically correct but lexically diverse~\cite{clipscore, pac-s, pac-spp}.

To address this limitation, recent work has proposed data-driven metrics that use pretrained vision--language models or multimodal encoders~\cite{vilbertscore, umic, pac-s}. 
Among them, Polos~\cite{polos} incorporates image information and supervised learning from human judgments. 
These metrics have been shown to align well with human judgments on standard image captioning evaluation benchmarks, most of which primarily contain short captions.
\begin{figure*}[t]
    \centering
    \includegraphics[width=0.9\linewidth]{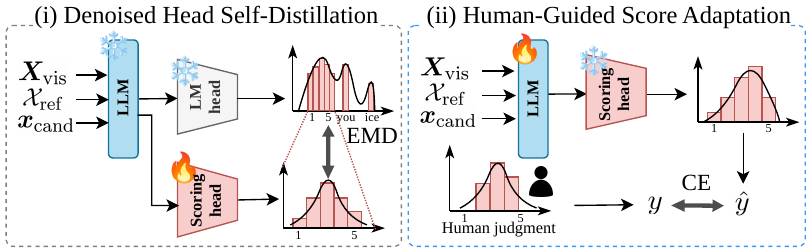}
    \caption{\textbf{Overview of our proposed two-phase training framework.} (i) Scoring head (red block) is trained with five labels using Earth Mover's Distance (EMD) while the LLM and the LM head are frozen. (ii) The LLM backbone is fine-tuned using human judgments while freezing the scoring head's parameters. CE represents cross-entropy.}
    \label{fig:model}
    \vspace{-1mm}
\end{figure*}
\paragraph{LLM-as-a-Judge approaches.}
Recent advances in LLMs and multimodal large language models (MLLMs) have led to a new family of evaluation metrics, often referred to as LLM-as-a-Judge approaches~\cite{survey-llm-judge}. 
These metrics evaluate captions using LLMs or MLLMs, often yielding more interpretable judgments than embedding-based similarity metrics. 
Several such approaches have been proposed for image captioning. 
For example, FLEUR~\cite{lee2024fleur} employs LLaVA~\cite{llava, llava-1.5} to incorporate image inputs directly for caption evaluation. 
Similarly, G-VEval~\cite{tong2025gveval} and HarmonicEval~\cite{ohi2024harmoniceval} score captions from multiple perspectives, improving interpretability and alignment with human evaluation criteria.

Despite their promising performance, LLM-based metrics suffer from a practical drawback: their scores are computed from LM-head logits over the full vocabulary, while caption evaluation only requires a small set of ordinal labels. As a result, much of the logit mass is assigned to task-irrelevant tokens.

\paragraph{Video captioning metrics.}
Compared with image captioning, automatic evaluation for video captioning has received less attention.
EMScore~\cite{emscore} extends embedding-based matching to the video domain;
PAC-S and PAC-S++~\cite{pac-s, pac-spp} compute CLIP-based similarity adapted for captioning evaluation;
FactVC~\cite{factvc} evaluates factual consistency between video and candidate captions; and
G-VEval~\cite{tong2025gveval} applies the LLM-as-a-Judge paradigm to video captioning.

\paragraph{Datasets and benchmarks.}
A variety of datasets have been used to evaluate image captioning metrics, including Composite~\cite{composite}, Flickr8K-Expert and Flickr8K-CF~\cite{flickr}, Polaris~\cite{polos}, and Nebula~\cite{deneb}. 
These benchmarks have played an important role in analyzing metric quality, but many of them provide human judgments from a single evaluation perspective, limiting their ability to assess fine-grained aspects of caption quality. 
This limitation is important for short captions, where minor wording differences can affect multiple quality dimensions simultaneously.

To provide fine-grained supervision, recent work has introduced datasets with multi-dimensional human judgments~\cite{ohi2024harmoniceval, thumb}. 
For video captioning, the VATEX-EVAL and ActivityNet-FOIL benchmarks~\cite{emscore}, as well as the ActivityNet-Fact and YouCook2-Fact benchmarks~\cite{factvc}, have been used to evaluate metrics.
However, they lack training sets with human judgments, making it difficult to develop supervised evaluation metrics. To address this limitation, we constructed Vid-Lepus (Section~\ref{sec:myvatex}), a dataset with human judgments for training supervised video captioning evaluation metrics.

\section{Method}
\label{sec:Rigel}

We propose \textsc{Rigel}, an automatic evaluation metric for both image and video captioning.
This metric improves LLM-based evaluation through self-distilled score adaptation framework, and it resolves the mismatch between large-vocabulary language modeling and evaluation tasks.
This scheme enables efficient evaluation with a single forward pass.
Fig.~\ref{fig:model} shows an overview of the proposed framework, which consists of two phases: (i) Denoised Head Self-Distillation and (ii) Human-Guided Score Adaptation. The proposed self-distilled score adaptation can be broadly applied to existing LLM-as-a-Judge frameworks.
Section~\ref{sec:phase1} describes Phase~1, which learns an evaluation-specific scoring head via self-distillation, Section~\ref{sec:phase2} describes Phase~2, which adapts the LLM backbone to human judgments, and Section~\ref{sec:inference} presents the inference procedure.

We define the input to the metric as a triplet $(\bm{X}_{\mathrm{vis}}, \mathcal{X}_{\mathrm{ref}}, \bm{x}_{\mathrm{cand}})$,
where
$\bm{X}_{\mathrm{vis}} \in \mathbb{R}^{T \times 3 \times H \times W}$ is a visual input consisting of $T$ frames of RGB images with height $H$ and width $W$ (with $T=1$ for still images),
$\mathcal{X}_{\mathrm{ref}} = \{ \bm{x}_{\mathrm{ref}}^{(i)} \}_{i=1}^N$ is a set of $N$ reference captions,
and $\bm{x}_{\mathrm{cand}}$ is a candidate caption.
In the reference-based setting, $\mathcal{X}_{\mathrm{ref}}$ consists of $N$ reference captions, whereas in the reference-free setting, $\mathcal{X}_{\mathrm{ref}} = \emptyset$. Given these inputs, the proposed metric outputs the final evaluation score $\hat{y}$.

\subsection{Phase~1: Denoised Head Self-Distillation}
\label{sec:phase1}

Phase~1 introduces an evaluation-specific scoring head distilled from a frozen LLM.
This head captures the model's judgment signals in a task-aligned label space via self-distillation.
We assume that the original LM head, which has been optimized for next-token prediction over a large vocabulary, is not well suited for predicting a small, predefined set of ordinal labels.
This mismatch can make the LM head unsuitable for direct use in evaluation.
To bridge this gap, we train a lightweight scoring head through self-distillation, transferring the LLM's evaluation capability from the noisy vocabulary space into a clean ordinal prediction space.

Given an input, we construct a prompt and process it with a frozen LLM to obtain a final-layer hidden representation $\bm{h} \in \mathbb{R}^{d}$, where $d$ denotes the hidden dimension.
From the vocabulary logits, we extract those corresponding to the ordinal score tokens and denote them by 
$\bm{z}_{\mathrm{score}}\in\mathbb{R}^{|\mathcal{M}|}$, where $\mathcal{M}$ denotes an ordinal label set; 
following previous studies \cite{polos, deneb,emscore}, we set $\mathcal{M}=\{1,2,3,4,5\}$.
We then apply a temperature-scaled softmax to obtain a soft pseudo-label distribution:

\begin{equation}
\bm{q} = \mathrm{softmax}(\bm{z}_{\mathrm{score}} / \tau),
\label{eq:pseudo}
\end{equation}
where $\tau$ is the distillation temperature.

We replace the LM head with a lightweight MLP scoring head $g_{\theta}\!:\mathbb{R}^{d}\!\rightarrow\!\mathbb{R}^{|\mathcal{M}|}$, which produces denoised logits:
$\hat{\bm{z}} = g_{\theta}(\bm{h}).$
We train $g_{\theta}$ by minimizing the Phase~1 distillation loss $\mathcal{L}_{\mathrm{P1}}$, defined as follows:
\begin{equation}
\mathcal{L}_{\mathrm{P1}} = \tau \, \mathrm{EMD}\!\left(\bm{q} , \mathrm{softmax}(\hat{\bm{z}} / \tau)\right),
\label{eq:loss_p1}
\end{equation}
where $\mathrm{EMD}$ denotes the one-dimensional Earth Mover's Distance (EMD)
over the ordered score labels~\cite{earth}, 
defined as follows:
\begin{equation}
    \mathrm{EMD}(\bm{a},\bm{b})
    =
    \sum_{i=1}^{K-1}
    \left|
    \sum_{j=1}^{i} a_j
    -
    \sum_{j=1}^{i} b_j
    \right|,
\end{equation}
where $\bm{a}$ and $\bm{b}$ denote probability distributions over $K$ ordered
labels, and $a_j$ and $b_j$ denote the probabilities assigned to the
$j$-th score label.
The factor $\tau$ is used to keep the gradient scale stable when applying temperature scaling.
During this phase, the LLM remains frozen and only the scoring head parameters $\theta$ are optimized.

\subsection{Phase~2: Human-Guided Score Adaptation}
\label{sec:phase2}

Phase~2 adapts the LLM backbone to human judgments under the frozen scoring head
$g_{\theta^{*}}$ learned in Phase~1, where $\theta^{*}$ denotes the scoring head parameters obtained in Phase~1. We keep $g_{\theta^{*}}$ frozen and fine-tune
only the backbone parameters via Low-Rank Adaptation (LoRA)~\cite{lora}, stabilizing training.
The backbone learns representations aligned with the scoring head.

Given a one-hot gold label $\bm{y}^{\mathrm{gold}} \in \{0,1\}^{|\mathcal{M}|}$ derived from averaged human judgment, we minimize the Phase~2 cross-entropy loss $\mathcal{L}_{\mathrm{P2}}$ as follows:
\vspace{-1mm}
\begin{equation}
\mathcal{L}_{\mathrm{P2}} = -\sum_{k=1}^{|\mathcal{M}|} y^{\mathrm{gold}}_{k} \log p_{k},
\label{eq:loss_p2}
\end{equation}
\vspace{-1mm}where $p_{k}$ denotes the $k$-th element of $\mathrm{softmax}(g_{\theta^{*}}(\bm{h}))$.
Only the LoRA parameters of the backbone are updated and $\theta^{*}$ remains fixed.

\subsection{Inference}
\label{sec:inference}

At inference, we compute $\hat{y}$ via score smoothing~\cite{lee2024fleur,tong2025gveval}:
$\hat{y} = \sum_{k=1}^{|\mathcal{M}|} k \cdot p_{k}$,
where $p_{k} = \mathrm{softmax}(g_{\theta^{*}}(\bm{h}))_{k}$.
We linearly normalize $\hat{y}$ to the range [0,1].

\begin{table*}[t]
\centering
\small
\begin{tabular}{lcccccccccc}
\toprule

\multirow{2}{*}{Methods} 
& \multicolumn{2}{c}{Flickr8K-Expert} 
& \multicolumn{2}{c}{Flickr8K-CF} 
& \multicolumn{2}{c}{Nebula} 
& \multicolumn{2}{c}{Composite} 
& \multicolumn{2}{c}{FOIL} \\

\cmidrule(lr){2-3} \cmidrule(lr){4-5} \cmidrule(lr){6-7} \cmidrule(lr){8-9} \cmidrule(lr){10-11}

& $\tau_b$ & $\tau_c$ 
& $\tau_b$ & $\tau_c$ 
& $\tau_b$ & $\tau_c$ 
& $\tau_b$ & $\tau_c$ 
& 1-ref[\%] & 4-ref[\%] \\

\midrule
\rowcolor{gray!15}
\multicolumn{11}{l}{\textbf{Reference-based}} \\

BLEU & 30.6 & 30.8 & 16.4 & 8.7 & 46.5 & 44.1 & 28.3 & 30.6 & 66.5 & 82.6 \\
ROUGE & 32.1 & 32.3 & 19.9 & 10.3 & 45.8 & 43.4 & 30.0 & 32.4 & 71.7 & 79.3 \\
CIDEr & 43.6 & 43.9 & 24.6 & 12.7 & 51.5 & 48.8 & 34.9 & 37.7 & 82.5 & 90.6 \\
METEOR & 41.5 & 41.8 & 22.2 & 11.5 & 50.2 & 47.6 & 36.0 & 38.9 & 78.8 & 82.6 \\
SPICE & 51.7 & 44.9 & 24.4 & 12.0 & 51.5 & 47.4 & 38.8 & 40.3 & 75.5 & 86.1 \\
BERTScore & 37.8 & 46.7 & 22.8 & 11.5 & 47.5 & 47.1 & 30.2 & 30.1 & 88.6 & 92.1 \\
RefCLIP-S & 52.6 & 53.0 & 36.4 & 18.8 & 53.6 & 50.8 & 51.2 & 55.4 & 91.0 & 92.6 \\
RefPAC-S & 55.5 & 55.9 & 37.6 & 19.5 & 54.7 & 51.9 & 53.0 & 57.3 & 93.7 & 94.9 \\
Polos & 56.1 & 56.4 & 37.8 & 19.5 & 58.0 & 55.0 & 53.7 & 57.6 & 93.3 & 95.4 \\
Ref-HICEScore & 57.2 & 57.7 & 38.2 & 19.8 & - & - & 53.9 & 58.7 & 96.4 & 97.0 \\
DENEB & 55.6 & 56.5 & 38.0 & 19.6 & 58.1 & 55.1 & 54.0 & 57.9 & 95.1 & 96.1 \\
RefPAC-S++ & 55.3 & 55.7 & 37.9 & 19.6 & 53.3 & 50.6 & 54.7 & 59.1 & 93.5 & 94.1 \\
Pearl & 58.2 & 58.6 & 38.6 & \underline{20.0} & \underline{58.4} & \underline{55.4} & \underline{55.8} & 60.4 & 96.5 & 97.2 \\

HiFiScore & - & 58.4 & - & - & - & - & - & \underline{65.8} & - & - \\
CLAIR & 58.3 & 48.8 & 38.2 & 17.0 & - & - & - & 61.0 & - & 93.6 \\
Ref-FLEUR & - & 51.9 & \underline{38.8} & - & - & - & - & 64.2 & 97.3 & \underline{98.4} \\

G-VEval & \textbf{60.5} & \underline{58.7} & 38.2 & 19.9 & - & - & - & - & \underline{97.8}  & \underline{98.4} \\

\rowcolor{blue!15}
Ours & \underline{58.6} & \textbf{59.0} & \textbf{40.4} & \textbf{20.9} & \textbf{60.6} & \textbf{57.5} & \textbf{61.2} & \textbf{66.1} & \textbf{99.1} & \textbf{99.2} \\

\midrule
\rowcolor{gray!15}
\multicolumn{11}{l}{\textbf{Reference-free}} \\

CLIP-S & 51.1 & 51.2 & 34.4 & 17.7 & 50.5 & 47.9 & 49.8 & 53.8 & 87.2 & 87.2 \\
PAC-S & 53.9 & 54.3 & 36.0 & 18.6 & 51.0 & 48.3 & 51.5 & 55.7 & 89.9 & 89.9 \\
BRIDGE & 55.4 & 55.8 & 36.3 & 19.0 & - & - & 52.9 & 57.2 & 93.0 & 93.0 \\
HICEScore & 55.9 & 56.4 & 37.2 & 19.2 & - & - & 53.1 & 57.9 & 93.1 & 93.1 \\
PAC-S++ & 54.1 & 54.5 & 37.0 & 19.1 & 50.5 & 47.9 & 53.9 & 58.3 & 90.2 & 90.2 \\
BLIP2-Score & 52.2 & 52.5 & 36.7 & 19.0 & 53.0 & 50.7 & 56.9 & 61.5 & 94.3 & 94.3 \\
Pearl & 56.2 & 56.6 & 37.8 & 19.5 & \underline{55.9} & 53.0 & 54.0 & 58.4 & 96.7 & 96.7 \\
HiFiScore & - & 58.4 & - & - & - & - & - & \underline{65.7} & - & - \\
FLEUR & - & 53.0 & 38.6 & - & - & - & - & 63.5 & 96.8 & 96.8 \\
G-VEval & \textbf{61.5} & \underline{59.7} & 38.7 & 20.2 & - & - & - & - & - & - \\
EXPERT & - & 56.7 & 39.3 & - & - & \underline{54.9} & - & 65.0 & - & - \\
DISCODE (LLaVA) & 55.7 & 56.1 & \underline{40.2} & \underline{20.8} & - & - & \textbf{61.1} & \textbf{66.0} & - & - \\
DISCODE (InternVL) & 57.7 & 58.1 & 40.1 & \underline{20.8} & - & - & \underline{60.5} & 64.9 & \underline{98.2} & \underline{98.2} \\
\rowcolor{blue!15}
Ours & \underline{59.5} & \textbf{59.9} & \textbf{40.4} & \textbf{20.9} & \textbf{58.1} & \textbf{55.1} & 59.5 & 64.3 & \textbf{98.7} & \textbf{98.7} \\

\bottomrule
\end{tabular}
\caption{\textbf{Quantitative comparison on image captioning evaluation benchmarks.} \textbf{Bold} font indicates the best results, and underlining indicates the second-best results. Following previous work~\cite{pearl, tong2025gveval, lee2024fleur}, we report the reproduced results for CLAIR. “-” indicates either non-executable code or unavailable data.}
\label{tab:image_results_full}
\end{table*}

\section{Vid-Lepus Dataset}
\label{sec:myvatex}
Supervised metrics for image and video captioning require a training dataset that includes candidates annotated with human judgments. Existing video-captioning evaluation datasets are insufficient because they do not provide training data with human judgments~\cite{factvc, emscore}.
To address this limitation, we constructed Vid-Lepus, a supervised video captioning dataset, for training evaluation metrics. This dataset comprises video clips, references, and candidate captions paired with human judgments.

From crowd workers, we collected human judgments on a five-point scale to evaluate the appropriateness of candidate captions with respect to the video clips and references.
Annotators were instructed to assess the candidates across three dimensions: descriptiveness, relevance, and fluency, following prior work~\cite{polos, vela}.
For quality control, we excluded annotations from evaluators exhibiting suspicious behavior, such as unusually short response times or consistently uniform ratings. Furthermore, samples whose human ratings exhibited a range of at least 3 on the five-point scale are manually reviewed by an expert annotator to resolve annotation disagreements.

To assess annotation reliability, we computed Krippendorff's $\alpha$ for inter-annotator agreement, which was 0.45. This value was higher than those of standard datasets, including CapEval1K (0.37)~\cite{umic} and Spica (0.44)~\cite{pearl}.
Further details on the dataset and its construction process are provided in Appendix \ref{sec:construction}.

\begin{table*}[t]
\centering
\small
\begin{tabular}{lccccccccccc}
\toprule

\multirow{3}{*}{Metrics} 
& \multicolumn{4}{c}{VATEX-EVAL} 
& \multicolumn{3}{c}{ActivityNet-Fact} 
& \multicolumn{3}{c}{YouCook2-Fact} 
& \multicolumn{1}{c}{ActivityNet-FOIL} \\

\cmidrule(lr){2-5} \cmidrule(lr){6-8} \cmidrule(lr){9-11}\cmidrule(lr){12-12} 

& \multicolumn{2}{c}{1-ref} 
& \multicolumn{2}{c}{9-ref} 
& Para & Sent & Word 
& Para & Sent & Word 
& Accuracy [\%] \\

\cmidrule(lr){2-3} \cmidrule(lr){4-5}  \cmidrule(lr){6-8} \cmidrule(lr){9-11}\cmidrule(lr){12-12}

& $\tau_b$ & $\rho$ & $\tau_b$ & $\rho$ 
& $r$ & $r$ & $r$ 
& $r$ & $r$ & $r$ & \\

\midrule

\rowcolor{gray!15}
\multicolumn{12}{l}{\textbf{Reference-based}} \\
EMScore 
& 28.6 & 37.1 & 36.8 & 47.2 
& 42.7 & 35.2 & 44.6 
& 54.3 & 51.4 & 55.3 
& 92.4 \\

RefPAC-S 
& 31.4 & 40.5 & 38.1 & 48.8 
& 47.0 & 37.8 & 49.5 
& 56.2 & 54.4 & 57.0 
& \underline{93.5} \\

RefPAC-S++ 
& 32.2 & \underline{41.5} & 39.8 & \underline{50.8} 
& - & - & - 
& - & - & - 
& 93.4 \\

FactVC 
& - & - & - & - 
& 55.1 & 46.5 & \underline{54.5} 
& 60.6 & 58.3 & 61.5 
& - \\

G-VEval 
& \underline{44.9} & - & \underline{48.1} & - 
& \underline{56.6} & \underline{48.1} & 48.8 
& \textbf{72.4} & \textbf{66.7} & \textbf{62.4} 
& - \\
\rowcolor{blue!15}
Ours
& \textbf{45.6} & \textbf{57.8} & \textbf{50.8} & \textbf{64.0} 
& \textbf{67.8} & \textbf{54.0} & \textbf{65.7} 
& \underline{65.4} & \underline{60.8} & \underline{61.7} 
& \textbf{97.1} \\

\midrule

\rowcolor{gray!15}
\multicolumn{12}{l}{\textbf{Reference-free}} \\
EMScore 
& 23.2 & 30.3 & 23.2 & 30.3 
& 25.3 & 19.0 & 30.0 
& 33.7 & 35.3 & 36.1 
& 89.5 \\

PAC-S 
& 25.1 & 32.6 & 25.1 & 32.6 
& 33.2 & 27.1 & 38.4 
& 31.2 & 33.5 & 33.1 
& 90.1 \\

PAC-S++ 
& 28.1 & \underline{36.4} & 28.1 & \underline{36.4} 
& - & - & - 
& - & - & - 
& 91.0 \\

FactVC 
& - & - & - & - 
& 46.2 & 37.1 & \underline{48.0} 
& 40.8 & 41.0 & 42.0 
& - \\

G-VEval 
& \underline{39.4} & - & \underline{39.4} & - 
& \underline{48.8} & \underline{38.6} & 43.6 
& \textbf{55.8} & \underline{50.1} & \underline{44.4} 
& - \\
\rowcolor{blue!15}
Ours
& \textbf{42.1} & \textbf{53.7} & \textbf{42.1} & \textbf{53.7} 
& \textbf{63.6} & \textbf{51.9} & \textbf{60.5} 
& \underline{55.0} & \textbf{50.3} & \textbf{50.0} 
& \textbf{96.2} \\
\bottomrule
\end{tabular}
\caption{\textbf{Quantitative comparison on video captioning evaluation benchmarks.} “$\rho$” and “$r$” represent Spearman’s and Pearson’s correlation coefficients, respectively. Bold indicates the best result and underlining indicates the second-best result in each column.  }
\label{tab:video_quantitative}
\end{table*}

\begin{table*}[t]
\centering
\small
\setlength{\tabcolsep}{3pt}
\renewcommand{\arraystretch}{1.08}

\begin{tabular}{c c c c c cc cc cc cc}
\toprule
\multirow{2}{*}{Metrics}
& \multirow{2}{*}{Phase 1}
& \multirow{2}{*}{Phase 2}
& \multirow{2}{*}{Head}
& \multirow{2}{*}{Backbone}
& \multicolumn{2}{c}{Composite}
& \multicolumn{2}{c}{Flickr8K-EX}
& \multicolumn{2}{c}{Nebula}
& \multicolumn{2}{c}{VATEX-Eval} \\
\cmidrule(lr){6-7}
\cmidrule(lr){8-9}
\cmidrule(lr){10-11}
\cmidrule(lr){12-13}
& & & &
& $\tau_b$ & $\tau_c$
& $\tau_b$ & $\tau_c$
& $\tau_b$ & $\tau_c$
& $\tau_b$ & $\rho$ \\
\midrule

(i)
&  & $\checkmark$ & LM head & Qwen3-VL-2B
& 58.3 & 63.0
& 59.0 & 59.4
& \underline{57.4} & \underline{54.4}
& 39.7 & 50.8 \\

(ii)
& $\checkmark$ &  & Scoring head & Qwen3-VL-2B
& 53.1 & 57.4
& 22.3 & 22.6
& 52.2 & 55.1
& \textbf{43.2} & \textbf{54.3} \\

(iii)
&  & $\checkmark$ & Scoring head & Qwen3-VL-2B
&  58.0 & 62.8
&  58.8 & 59.2
&  56.3 & 53.4
&  35.4 & 45.4 \\

(iv)
& $\checkmark$ & $\checkmark$ & Scoring head & Qwen2.5-VL-3B
&  \underline{59.4}  & \underline{64.2}
&  \underline{59.2}   & \underline{59.7}
& 57.3 & 54.3
& 39.8 & 50.8 \\

\rowcolor{blue!15}
(v)
& $\checkmark$ & $\checkmark$ & Scoring head & Qwen3-VL-2B
& \textbf{59.5} & \textbf{64.3} & \textbf{59.5} & \textbf{59.9} & \textbf{58.1} & \textbf{55.1} & \underline{42.1} & \underline{53.7} \\

\bottomrule
\end{tabular}

\caption{\textbf{Comparison across different training phases, heads, and backbones in the reference-free setting.} \textbf{Bold} indicates the best value in each column, and underlining indicates the second-best value. These results demonstrate that both Phases 1 and 2 contributed to the performance improvement on most benchmarks.}
\label{tab:ablation} 
\vspace{-2mm}
\end{table*}

\section{Experiments}
\label{sec:experiments}
\subsection{Experimental Setup}
\paragraph{Datasets.}

For the image captioning evaluation, we used the standard benchmarks Flickr8K-Expert~\cite{flickr}, Flickr8K-CF~\cite{flickr}, Nebula~\cite{deneb}, Composite~\cite{composite}, and FOIL~\cite{foil}. 
To evaluate the video captioning metric, we used VATEX-EVAL~\cite{emscore}, ActivityNet-FOIL~\cite{emscore}, ActivityNet-Fact~\cite{factvc}, and YouCook2-Fact~\cite{factvc}.
To train \textsc{Rigel}, we used the constructed Vid-Lepus dataset for video captioning along with Spica~\cite{pearl} for image captioning.
The Spica dataset~\cite{pearl} was used with its original training and validation splits, containing $296,149$ and $3,000$ samples, respectively. 
The Vid-Lepus dataset was divided into training and validation sets containing $13,262$ and $1,540$ samples, respectively.
The training and validation sets of the datasets were used for metric training and hyperparameter tuning, respectively.
Details of the datasets are provided in Appendix \ref{sec:construction}.
\begin{figure*}[t]
    \centering
    \includegraphics[width=1.0\linewidth]{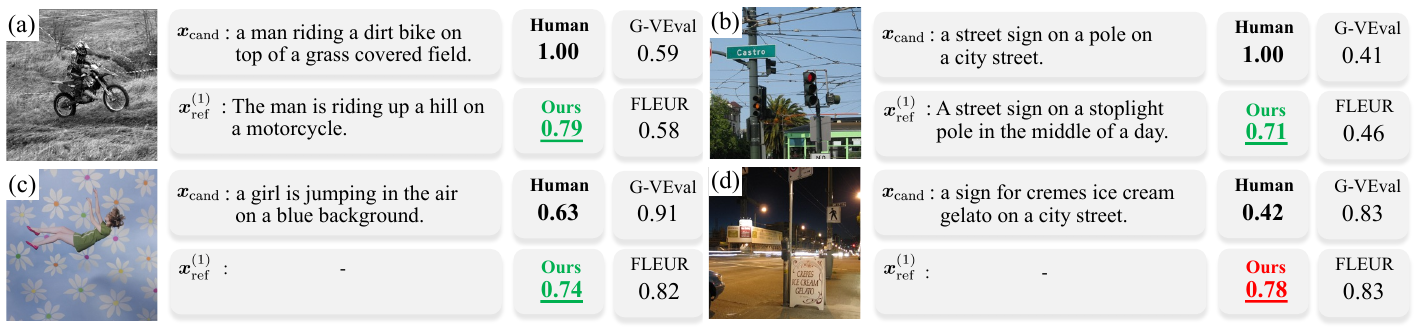}
    \caption{\textbf{Qualitative results on the Nebula dataset.} Cases (a)–(b) illustrate successful examples in the reference-based setting, whereas (c) shows a successful example in the reference-free setting. In contrast, (d) represents a failure case in the reference-free setting. \textcolor{mygreen}{Green} values indicate predictions closest to human annotations, and \textcolor{red}{red} values denote critical errors. “-” indicates that no reference caption was provided.}
    \label{fig:qualitative-image}
    \vspace{-3mm}
\end{figure*}

\paragraph{Baselines.}

We selected multiple standard metrics for captioning evaluation as our baseline comparison metrics.
For both image and video captioning, we used embedding-based metrics PAC-S~\cite{pac-s} and PAC-S++~\cite{pac-spp}; and the LLM-as-a-Judge metric G-VEval~\cite{tong2025gveval}.
For image captioning, we additionally included traditional reference-based lexical matching metrics, namely BLEU~\cite{bleu}, ROUGE~\cite{rouge}, METEOR~\cite{meteor}, CIDEr~\cite{cider}, and SPICE~\cite{spice}.
We also included representative embedding-based metrics, including CLIPScore~\cite{clipscore}, BERTScore~\cite{bertscore}, HICEScore~\cite{Zeng_2024hicescore}, BRIDGE~\cite{bridge}, and BLIP2-Score~\cite{blip2score}; supervised metrics trained on human judgments, including Polos~\cite{polos}, DENEB~\cite{deneb}, and Pearl~\cite{pearl}; and LLM-as-a-Judge metrics, including CLAIR~\cite{chan2023clair}, FLEUR~\cite{lee2024fleur}, HiFiScore~\cite{yao2024hifi}, EXPERT~\cite{expert}, and DISCODE~\cite{discode}. For reproducibility, we used CaptionEvalKit-for-VLMs\footnote{\scriptsize{
\url{https://github.com/YuigaWada/CaptionEvalKit-for-VLMs}}} to assess these metrics.
For video captioning, we further adopted two representative metrics in this field: EMScore~\cite{emscore} and FactVC~\cite{factvc}. All experiments were reported based on a single run.

\paragraph{Evaluation metrics.}
We followed the standard evaluation practice used for each benchmark. Specifically, we used Kendall's $\tau_b$ and $\tau_c$ for the Composite, Flickr8K-Expert, Flickr8K-CF, and Nebula datasets; accuracy for FOIL and ActivityNet-FOIL; Pearson's correlation coefficient $r$ for ActivityNet-Fact and YouCook2-Fact; and Kendall's $\tau_b$ together with Spearman's $\rho$ for VATEX-EVAL.
\vspace{-3mm}
\subsection{Quantitative Results}
\vspace{-1mm}
\paragraph{Image captioning evaluation.}

Table~\ref{tab:image_results_full} presents a quantitative comparison with baseline metrics on the Composite, Flickr8K-Expert, Flickr8K-CF, Nebula, and FOIL datasets.
In the reference-based setting, our metric achieved strong performance across most benchmarks.
On the Composite dataset, our metric outperformed the previous best baseline, and on FOIL, our metric achieved the highest accuracy in both the 1-ref and 4-ref settings.
In the reference-free setting, our metric yielded competitive results, outperforming the baselines on several benchmarks including Flickr8K-CF, Nebula, and FOIL.
These results demonstrate that our self-distilled score adaptation effectively improves the correlation with human judgments across image captioning benchmarks.

\paragraph{Video captioning evaluation.}

Table~\ref{tab:video_quantitative} presents a quantitative comparison of metrics on VATEX-EVAL, ActivityNet-Fact, YouCook2-Fact, and ActivityNet-FOIL.
On VATEX-EVAL, our metric in the reference-based setting achieved improvements over the baselines in both $\tau_b$ and $\rho$ across all reference settings (1-ref, 9-ref).
On ActivityNet-FOIL, our metric achieved the highest accuracy, outperforming the best baseline score by a substantial margin.
Table~\ref{tab:video_quantitative} shows that our proposed metric achieved the best Pearson's $r$ on ActivityNet-Fact at the paragraph, sentence, and word levels.
Similarly, in the reference-free setting, our metric outperformed all baseline metrics across most benchmarks, confirming the effectiveness of our approach for video captioning evaluation.

\subsection{Ablation Study}
\label{sec:ablation}

\begin{figure*}[t]
    \centering
    \includegraphics[width=1.0\linewidth]{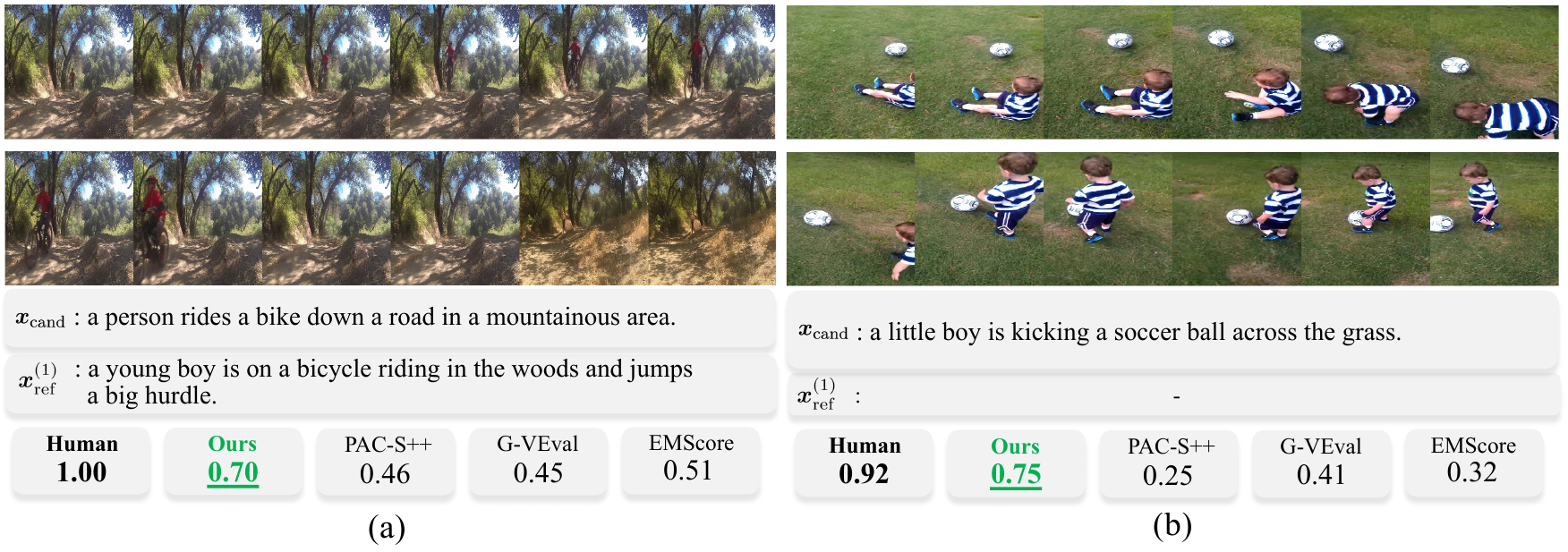}
    \vspace{-3mm}
    \caption{\textbf{Examples of successful cases from the VATEX-EVAL dataset.} Case (a) illustrates a successful example in the reference-based setting, whereas (b) shows a successful example in the reference-free setting. }
    \label{fig:qualitative-video}
    \vspace{-4mm}
\end{figure*}

Table \ref{tab:ablation} shows the quantitative results of the ablation studies.
We conducted three ablation studies to investigate the contribution of each component in our proposed metric: Phase 1, Phase 2, and the backbone model.

\paragraph{Denoised Head Self-Distillation Ablation.}
We investigated the contribution of Phase 1 by excluding it from the full pipeline.
As shown in Table \ref{tab:ablation}, a comparison between Metric (iii), which has a randomly initialized scoring head, and Metric (v) indicates that excluding Phase 1 decreased the scores on all benchmarks.
Specifically, Metric (v) outperformed Metric (iii) by 1.5 points on Composite, 0.7 points on Flickr8K-EX, and 1.8 and 1.7 points on Nebula for $\tau_b$ and $\tau_c$, respectively. On VATEX-Eval, Metric (v) outperformed Metric (iii) by 6.7 and 8.3 points for $\tau_b$ and $\rho$, respectively.
These results demonstrate that Phase 1 consistently improves performance. Furthermore, we compared Metrics (i) and (v) to assess the effect of using a scoring head within the full pipeline instead of the LM head. The results indicate that the scoring head is more effective than the LM head across all benchmarks.
Specifically, Metric (v) outperformed Metric (i) by 1.2 and 1.3 points on Composite, 0.5 and 0.5 points on Flickr8K-EX, 0.7 and 0.7 points on Nebula for $\tau_b$ and $\tau_c$, respectively.
\paragraph{Human-Guided Score Adaptation Ablation.}
We investigated the contribution of Phase 2 by excluding it from the full pipeline.
As shown in Table \ref{tab:ablation}, a comparison between Metrics (ii) and (v) indicates that excluding Phase 2 degrades the performance.
Specifically, Metric (v) outperformed Metric (ii) by 6.4 and 6.9 points on Composite, 37.2 and 37.3 points on Flickr8K-EX for $\tau_b$ and $\tau_c$, respectively.
These results demonstrate that Phase 2 substantially improves performance. 
\paragraph{Backbone Ablation.}
We investigated the effect of the backbone model by replacing Qwen3-VL-2B~\cite{qwen3vl} with Qwen2.5-VL-3B~\cite{qwen25vl}.
As shown in Table \ref{tab:ablation}, Metric (v) outperformed Metric (iv) by 0.3 and 0.2 points on Flickr8K-Expert, 0.8 and 0.8 points on Nebula for $\tau_b$ and $\tau_c$, and 2.3 and 2.9 points on VATEX-Eval for $\tau_b$ and $\rho$.
These results indicate that Qwen3-VL-2B performs at least as well as Qwen2.5-VL-3B and improves the results on Flickr8K-Expert, Nebula, and VATEX-Eval.

\subsection{Qualitative Results}

\paragraph{Image captioning.}
Fig. \ref{fig:qualitative-image} presents representative examples of the results of the proposed metric on the Nebula dataset. We used Nebula for the qualitative analysis because it is a diverse and balanced dataset~\cite{deneb}.
Cases (a) and (b) show successful cases in the reference-based setting, while Case (c) displays a successful case in the reference-free setting. Case (d) illustrates a sample for which the proposed metric did not perform as expected. 

In Case (a), $\bm{x}_{\text{ref}}^{(1)}$ was ``the man is riding up a hill on a motorcycle.'' while $x_{\text{cand}}$ was ``a man riding a dirt bike on top of a grass covered field.'' In this sample, the human judgment was $1.00$ as $x_{\text{cand}}$ appropriately describes the image. However, G-VEval and FLEUR evaluated it as $0.59$ and $0.58$, respectively, while the proposed metric rated it as $0.79$. Similarly, in Case (b), the metric yielded the score most aligned with human judgment.

In Case (c), $\bm{x}_{\mathrm{cand}}$ was "a girl is jumping in the air on a blue background." For this caption, the average human judgment score was moderate at 0.63, because the caption mischaracterizes the girl as jumping. While G-VEval, FLEUR, and the original Qwen3-VL-2B using its LM head (Original LLM) provided scores of 0.91, 0.82, and 0.41, respectively, the proposed metric evaluated it at 0.74, closely aligned with the human judgment.

In Case (d), $\bm{x}_{\mathrm{cand}}$ contains critical errors: it misread the text on the sign, rendering "crepes" as "cremes" and also failed to mention the background. Therefore, the average human judgment score was moderate at 0.42.
However, our proposed metric gave a relatively high score of 0.78, indicating a discrepancy with respect to human judgment. Similarly, G-VEval, FLEUR, and the original LLM output scores of 0.83, 0.83, and 0.72, respectively. These results indicate that these metrics may
not adequately penalize textual errors in scene text.

\paragraph{Video captioning.}
Fig. \ref{fig:qualitative-video} presents qualitative results from experiments on the VATEX-EVAL dataset. Cases (a) and (b) illustrate successful cases in the reference-based and reference-free settings.  In Case (a), $\bm{x}_{\text{ref}}^{(1)}$ was ``a young boy is on a bicycle riding in the woods and jumps
a big hurdle.'' and $x_{\text{cand}}$ was ``a person rides a bike down a road in a mountainous area.” The human judgment was $1.00$ because $x_{\text{cand}}$ appropriately describes the video. However, PAC-S++, G-VEval, and EMScore evaluated it as $0.46$, $0.45$ and $0.51$, respectively, while the proposed metric rated it as $0.70$.

Case (b) is a sample rated as moderately good by human evaluators with a score of $0.92$. This score reflects a partially correct description that was semantically aligned with the video but omitted some visual details.
The proposed metric rated it as $0.75$, while PAC-S++ underestimated it as $0.25$, G-VEval as $0.41$, and EMScore as $0.32$.

\section{Conclusion}
In this study, we addressed the task of automatic evaluation for image and video captioning and proposed \textsc{Rigel}, a unified automatic evaluation metric.
This metric improves LLM-based evaluation through self-distilled score adaptation framework.
It resolves the mismatch between large-vocabulary language modeling and evaluation tasks that require prediction over a small set of ordinal labels.
We first introduced an evaluation-specific scoring head distilled from a frozen LLM.
We then refined the LLM backbone using human judgment data while freezing the head's parameters.
Furthermore, we constructed Vid-Lepus, a new dataset for training human-aligned metrics for video captioning.
Experiments on multiple benchmarks showed that the method achieves higher correlation with human judgments than existing methods.

\section{Limitations}
Although our metric achieves a high correlation with human judgments, it has the following limitations.
First, Phase~1 relies on the soft pseudo-labels extracted from the frozen LM head.
Improving the self-distillation procedure could further enhance the quality of the scoring head.
Second, Phase~2 adapts the backbone using standard LoRA fine-tuning.
However, developing adaptation methods specifically designed for evaluation tasks could yield additional gains. Third, our metric requires access to the hidden representations of the underlying LLM, which prevents the direct use of proprietary models.

\section*{Acknowledgments}
This work was supported by funding from Apple Inc. Any views, opinions, findings, and conclusions or recommendations expressed in this material are those of the authors and should not be interpreted as reflecting the views, policies, or position, either expressed or implied, of Apple Inc.
This work was also partially supported by JSPS KAKENHI
Grant Number 23K28168, JST Moonshot, and JSPS Fellows Grant Number JP25KJ2069.

\bibliography{reference}

\appendix
\vspace{-2mm}
\begin{figure}[h]
    \centering
    \includegraphics[width=0.9\linewidth]{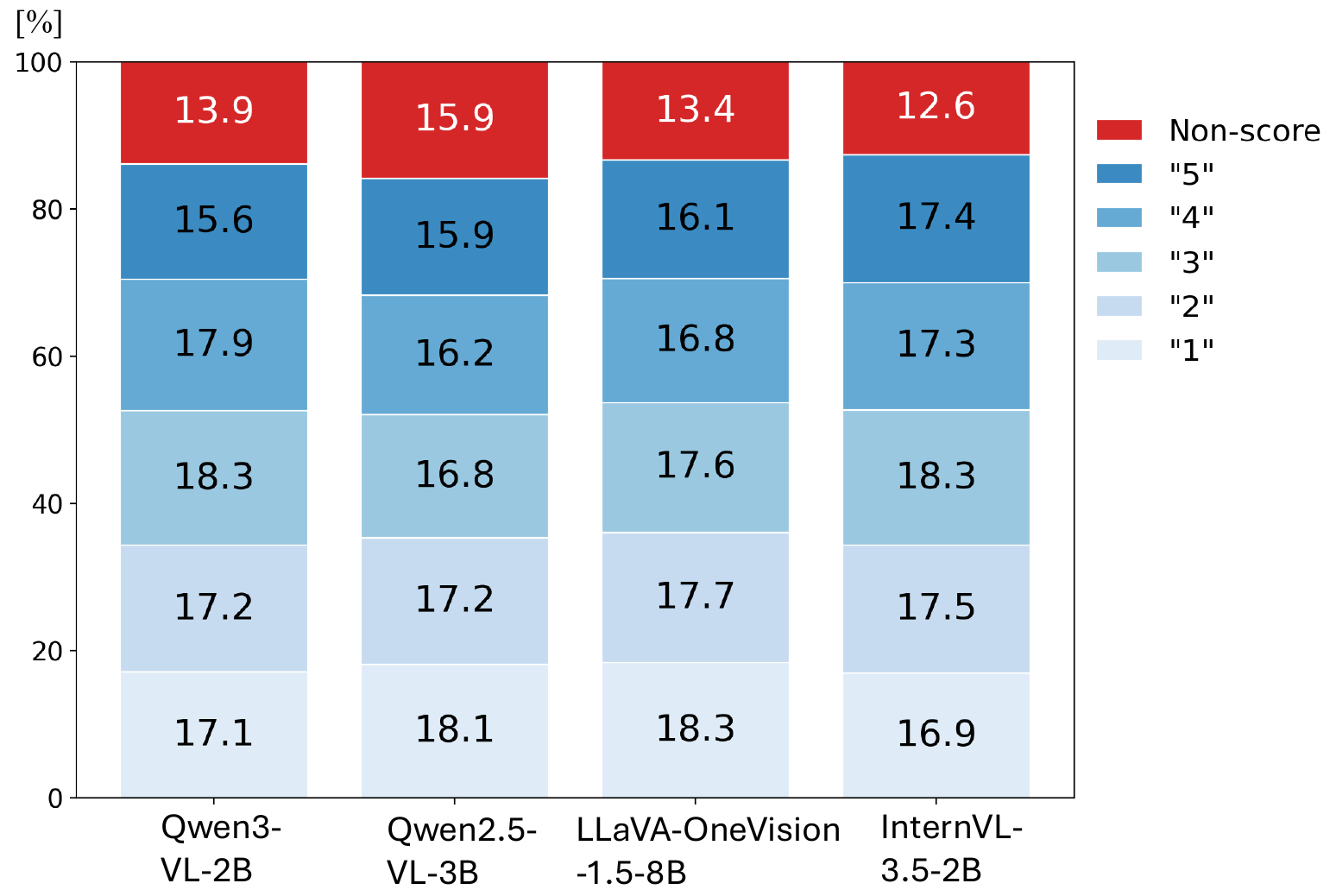}
   \caption{The logit distribution over score tokens (``1''–``5'') and non-score tokens on the Composite dataset~\cite{composite}. Non-score tokens exhibit logit magnitudes comparable to those of score tokens.
   }
    \label{fig:token_logits_composite}
    \vspace{-3mm}
\end{figure}

\section{Additional Related Work}
\label{sec:additional_related_work}

\begin{table*}
\centering
\begin{tabular}{lcc}
\toprule
 & \textbf{Phase 1} & \textbf{Phase 2} \\
\midrule
Epoch 
    & 50 
    & 1 (image) / 1 (video) \\

Optimizer 
    & Adam ($\beta_1=0.9$, $\beta_2=0.999$) 
    & Adam ($\beta_1=0.9$, $\beta_2=0.999$) \\

Learning rate 
    & $5 \times 10^{-4}$ 
    & $1 \times 10^{-4}$ \\

Batch size 
    & 64 
    & 8 \\

Weight decay 
    & 0.01 
    & 0.01 \\

LoRA rank 
    & -- 
    & 8 \\

LoRA alpha 
    & -- 
    & 16 \\

\bottomrule
\end{tabular}
\caption{Settings of the proposed metric for Phases 1 and 2.}
\label{tab:settings}
\end{table*}

\paragraph{Image captioning metrics.}
Automatic evaluation metrics for image captioning such as BLEU~\cite{bleu}, METEOR~\cite{meteor}, ROUGE~\cite{rouge}, CIDEr~\cite{cider}, and SPICE~\cite{spice} have traditionally relied on reference-based lexical or semantic matching. 
Several extensions, such as CIDEr-R~\cite{cider-r} and JaSPICE~\cite{jaspice}, have also been proposed to improve robustness or adapt evaluation to specific settings. 
Although these metrics remain standard in the literature, prior studies have shown that they often correlate only weakly with human judgments, especially when captions are semantically correct but lexically diverse~\cite{clipscore, pac-s, pac-spp}.

To address this limitation, recent work has proposed data-driven metrics that leverage pretrained vision--language models or multimodal encoders~\cite{vilbertscore, umic, clipscore, pac-s, pac-spp, polos, deneb}. 
Among them, CLIPScore~\cite{clipscore} evaluates captions by measuring image--text similarity in a reference-free manner, while PAC-S and PAC-S++~\cite{pac-s, pac-spp} improve this paradigm by adapting CLIP-based scoring to image caption evaluation.  
Other approaches, such as ViLBERTScore~\cite{vilbertscore}, UMIC~\cite{umic}, Polos~\cite{polos}, and DENEB~\cite{deneb}, further incorporate image information and/or supervised learning from human judgments. 
These metrics have shown strong performance on standard image captioning benchmarks, most of which primarily consist of short captions.

\section{Logit Distribution Analysis}
\label{additional_logit_distribution}
\vspace{-3mm}

We investigated the logit distributions over score tokens (“1”–“5”) and non-score tokens as follows: For each sample in the dataset, we extracted the LM-head logits corresponding to the score tokens and aggregated the logits of all remaining vocabulary tokens into a single non-score logit using the log-sum-exp operator. We then normalized these six quantities within each sample and averaged them across samples.

Fig.~\ref{fig:token_logits_composite} shows the distribution of LM-head logits assigned to score tokens (``1''--``5'') and the remaining vocabulary on the Composite dataset. Non-score tokens exhibit logit magnitudes comparable to those of the score tokens across the four models (Qwen3-VL-2B~\cite{qwen3vl}, Qwen2.5-VL-3B~\cite{qwen25vl}, LLaVA-OneVision-1.5-8B~\cite{llavaov1.5}, and InternVL-3.5-2B~\cite{internvl35}). This observation provides partial evidence for our claim that non-score tokens function as noise in score prediction.

\section{Distributional Objectives for Phase~1 Distillation}
\label{app:phase1_objective}
\label{app:emd}
We use EMD for Phase~1 distillation because the
teacher distribution is defined over the ordinal five-point score space.
Given two probability distributions
$\bm{a}, \bm{b}$ over $K$ ordered labels, one-dimensional EMD~\cite{earth} is defined as follows:
\begin{equation}
    \mathrm{EMD}(\bm{a},\bm{b})
    =
    \sum_{i=1}^{K-1}
    \left|
    \sum_{j=1}^{i} a_j
    -
    \sum_{j=1}^{i} b_j
    \right|,
\end{equation}
where $a_j$ and $b_j$ denote the probabilities assigned to the $j$-th score
label.
At inference, the predicted distribution is converted into a scalar score by
taking the expectation over ordered labels, making label distances relevant to
Phase~1 distillation.
KL divergence is insufficient for this purpose because it compares the
probabilities assigned to individual labels without incorporating distances in
the ordinal five-point score space.
In contrast, EMD penalizes discrepancies according to their distance along the
ordered label axis, aligning the training objective with the expectation-based
inference procedure.
\begin{table}[t]
\centering
\small
\setlength{\tabcolsep}{6pt}
\renewcommand{\arraystretch}{1.08}
\begin{tabular}{lcc}
\toprule
Caption properties& $r$ & $\rho$ \\
\midrule
Caption length          & 0.034  & 0.024 \\
Vocabulary uniqueness   & -0.017 & 0.010 \\
Descriptiveness density & 0.068  & 0.090 \\
\bottomrule
\end{tabular}
\caption{Correlations between human judgments and caption properties on Vid-Lepus.}
\label{tab:caption_properties}
\end{table}

\begin{table*}[t]
\centering
\small
\setlength{\tabcolsep}{5pt}
\renewcommand{\arraystretch}{1.08}
\begin{tabular}{lccccccc}
\toprule
\multirow{2}{*}{Methods} 
& \multirow{2}{*}{Prompt}
& \multicolumn{1}{c}{Composite}
& \multicolumn{1}{c}{Flickr8K-EX}
& \multicolumn{1}{c}{Flickr8K-CF}
& \multicolumn{1}{c}{Nebula}
& \multicolumn{2}{c}{VATEX-EVAL} \\
\cmidrule(lr){3-3}
\cmidrule(lr){4-4}
\cmidrule(lr){5-5}
\cmidrule(lr){6-6}
\cmidrule(lr){7-8}
&
& $\tau_c$
& $\tau_c$
& $\tau_c$
& $\tau_c$
& $\tau_b$
& $\rho$ \\
\midrule
\textsc{Rigel}
& G-VEval
& \textbf{65.6}
& 58.9
& \textbf{20.9}
& \textbf{56.4}
& \textbf{42.4}
& \textbf{54.0} \\

\rowcolor{blue!15}
\textsc{Rigel}
& Original
& 64.3
& \textbf{59.9}
& \textbf{20.9}
& 55.1
& 42.1
& 53.7 \\
\bottomrule
\end{tabular}
\caption{Comparison of \textsc{Rigel} with the G-VEval prompt and the original prompt.}
\label{tab:rigel_prompt_comparison}
\end{table*}
\begin{table*}[t]
\centering
\small
\setlength{\tabcolsep}{5pt}
\renewcommand{\arraystretch}{1.08}
\begin{tabular}{lccccccc}
\toprule
\multirow{2}{*}{Methods}
& \multirow{2}{*}{Backbone}
& \multicolumn{1}{c}{Composite}
& \multicolumn{1}{c}{Flickr8K-EX}
& \multicolumn{1}{c}{Flickr8K-CF}
& \multicolumn{1}{c}{Nebula}
& \multicolumn{2}{c}{VATEX-EVAL} \\
\cmidrule(lr){3-3}
\cmidrule(lr){4-4}
\cmidrule(lr){5-5}
\cmidrule(lr){6-6}
\cmidrule(lr){7-8}
&
& $\tau_c$
& $\tau_c$
& $\tau_c$
& $\tau_c$
& $\tau_b$
& $\rho$ \\
\midrule
G-VEval
& Qwen3-VL-2B
& 53.1
& 31.5
& 14.8
& 46.2
& 34.0
& 44.1 \\

FLEUR
& Qwen3-VL-2B
& 60.3
& 46.0
& 18.4
& 51.2
& --
& -- \\

\rowcolor{blue!15}
\textsc{Rigel} (Original)
& Qwen3-VL-2B
& \textbf{64.3}
& \textbf{59.9}
& \textbf{20.9}
& \textbf{55.1}
& \textbf{42.1}
& \textbf{53.7} \\
\bottomrule
\end{tabular}
\caption{Comparison with existing caption evaluation methods using the same backbone.}
\label{tab:same_backbone_comparison}
\end{table*}
\section{Construction of Vid-Lepus}
\vspace{-3mm}
\label{sec:construction}
\paragraph{Statistics.}
The Vid-Lepus dataset consists of $3,338$ video clips extracted from the validation split of the VATEX dataset~\cite{emscore}. 
Each video is paired with $10$ fixed reference captions, and the number of candidate captions per video ranges from 1 to 6, with an average of $1.69$. 
In total, the dataset contains $5,637$ candidate captions, with a vocabulary size of $7,029$ words, a total word count of $134,960$, and an average length of $23.9$ words. 
The reference set includes $33,380$ captions, with a vocabulary size of $11,030$ words, a total word count of $804,000$, and an average length of $14.3$ words. 
All captions are in English. The dataset also includes $14,802$ human judgments collected from $241$ annotators, which is an average of 2.63 judgments per candidate caption.
We confirmed that there is no overlap in videos or captions
between Vid-Lepus and VATEX-EVAL, based on MD5 checksums of all video files
and exact caption matching.
\paragraph{Annotation process.}
We recruited annotators from a general population on the internet via a public crowdsourcing platform, without restricting demographic or geographic background.
Annotators were recruited and compensated appropriately based on their country of residence, and consent was obtained through the task instructions, which clearly stated that the collected data would be used for research purposes.

Annotators were instructed to assess the appropriateness of candidate captions
with respect to the given videos across three dimensions: descriptiveness,
relevance, and fluency, each scored on a five-point scale.
The scoring criteria were as follows:
\begin{itemize}
    \setlength{\parskip}{0.5mm}
    \setlength{\itemsep}{0.2mm}
    \item[5:] \underline{Excellent} — The caption comprehensively and
    accurately describes all observed objects, relationships, and contextual details,
    with no grammatical errors or at most one minor error.
    \item[4:] \underline{Good} — The caption describes most objects and
    relationships with only minor omissions or inaccuracies, and is generally
    natural and comprehensible.
    \item[3:] \underline{Fair} — The caption mentions key objects
    but lacks detail in relationships or other attributes, or contains significant
    inaccuracies or noticeable grammatical errors, yet remains understandable.
    \item[2:] \underline{Poor} — The caption includes descriptions of a few objects
    but omits significant details, contains numerous inaccuracies, or has errors
    that make it difficult to read.
    \item[1:] \underline{Bad} — The caption provides minimal description,
    is fundamentally unrelated to the video content, or contains frequent errors
    that render it incomprehensible.
\end{itemize}

For quality control, we excluded annotations from evaluators exhibiting suspicious behavior, such as unusually short response times or consistently uniform ratings. Furthermore, samples whose human ratings exhibited a range of at least 3 on the five-point scale are manually reviewed by an expert annotator to resolve annotation disagreements. The five-point human judgment scores were normalized to the range $[0,1]$.

\paragraph{Annotation bias analysis.}
To investigate whether annotators systematically prioritize certain sentence structures, we conducted an additional analysis. 
Specifically, we computed correlations between the human judgments and three properties characterizing sentence structures on Vid-Lepus, defined in previous works~\cite{captionsmiths}: caption length, vocabulary uniqueness, and descriptiveness density.

Table \ref{tab:caption_properties} presents the correlation coefficients between the human judgments and the three properties. The table shows that caption length yielded 0.034 and 0.024 for Pearson's r and Spearman's $\rho$, respectively, vocabulary uniqueness yielded $-0.017$ and 0.010, and descriptiveness density yielded 0.068 and 0.090. These near-zero correlations indicate that annotators do not systematically prioritize particular sentence structures.

\section{Implementation Details}

Table~\ref{tab:settings} shows the settings of the proposed metric for Phases 1 and 2.
In Phase 1, the scoring head had 2.63M trainable parameters and an inference cost of 0.01 GFLOPs. 
In Phase 2, the model had 1.61M trainable parameters and required an average of 770 GFLOPs per inference. We trained the proposed metric on eight NVIDIA H200 SXM GPUs (141\,GB VRAM per GPU) and performed evaluation on a single H200 GPU. 
The total training time was approximately $11.6$ hours, and the average inference time was approximately 610\,ms per sample.
In Phase 1, we employed early stopping based on Kendall’s $\tau_c$. 
Specifically, $\tau_c$ was computed on the validation set after each epoch, and training was stopped when no improvement in $\tau_c$ was observed for five consecutive epochs. 
The model achieving the highest $\tau_c$ was then evaluated on the test sets. 
In Phase 2, the model was trained for one epoch.

\section{Additional Experiments} 
\subsection{Effect of Prompt Template}
We conducted controlled experiments to disentangle the effect of the prompt template. Specifically, we replaced \textsc{Rigel}’s original prompt template with the prompt used by G-VEval.  Table \ref{tab:rigel_prompt_comparison} shows the comparison with this variant. Table \ref{tab:rigel_prompt_comparison} shows that \textsc{Rigel} with the G-VEval prompt achieved performance comparable to the original \textsc{Rigel} and even outperformed it on several benchmarks. This result suggests that prompt engineering could further improve the proposed metric.
\begin{table}[t]
\centering
\small
\setlength{\tabcolsep}{4pt}
\renewcommand{\arraystretch}{1.08}
\resizebox{\columnwidth}{!}{
\begin{tabular}{lcccc}
\toprule
\multirow{2}{*}{Methods}
& \multirow{2}{*}{\shortstack{Training dataset\\for image captioning}}
& \multicolumn{1}{c}{Composite}
& \multicolumn{1}{c}{Flickr8K-EX}
& \multicolumn{1}{c}{Flickr8K-CF} \\
\cmidrule(lr){3-3}
\cmidrule(lr){4-4}
\cmidrule(lr){5-5}
&
& $\tau_c$
& $\tau_c$
& $\tau_b$ \\
\midrule
Pearl
& Spica
& 60.4
& 58.6
& 38.6 \\

DENEB
& Spica
& 58.6
& 56.9
& 38.2 \\

\rowcolor{blue!15}
Ours
& Spica
& \textbf{66.1}
& \textbf{59.0}
& \textbf{40.4} \\
\bottomrule
\end{tabular}
}
\caption{Quantitative comparison with same-data baselines on image captioning evaluation benchmarks.}
\label{tab:same_data_baselines}
\end{table}
\begin{table}[t]
\centering
\small
\setlength{\tabcolsep}{4pt}
\renewcommand{\arraystretch}{1.08}
\resizebox{\columnwidth}{!}{
\begin{tabular}{lccc}
\toprule
\multirow{2}{*}{Methods}
& \multirow{2}{*}{\shortstack{Training dataset\\for video captioning}}
& \multicolumn{2}{c}{VATEX-EVAL} \\
\cmidrule(lr){3-4}
&
& $\tau_b$
& $\rho$ \\
\midrule
PAC-S++
& Vid-Lepus
& 31.2
& 40.3 \\

EMScore 
& Vid-Lepus
& 30.2
& 39.0 \\

\rowcolor{blue!15}
Ours
& Vid-Lepus
& \textbf{42.1}
& \textbf{53.7} \\
\bottomrule
\end{tabular}
}
\caption{Quantitative comparison with same-data baselines on VATEX-EVAL.}
\label{tab:same_data_baselines_vatex}
\end{table}
\begin{table}[t]
\centering
\small
\setlength{\tabcolsep}{4pt}
\renewcommand{\arraystretch}{1.08}

\resizebox{\columnwidth}{!}{
\begin{tabular}{c c cc cc cc cc}
\toprule
\multirow{2}{*}{Metrics}
& \multirow{2}{*}{Objective}
& \multicolumn{2}{c}{Composite}
& \multicolumn{2}{c}{Flickr8K-EX}
& \multicolumn{2}{c}{Nebula}
& \multicolumn{2}{c}{VATEX-Eval} \\
\cmidrule(lr){3-4}
\cmidrule(lr){5-6}
\cmidrule(lr){7-8}
\cmidrule(lr){9-10}
&
& $\tau_b$ & $\tau_c$
& $\tau_b$ & $\tau_c$
& $\tau_b$ & $\tau_c$
& $\tau_b$ & $\rho$ \\
\midrule

(i)
& KL
& 59.0 & 63.8
& 59.3 & 59.7
& 57.7 & 54.7
& 41.9 & 53.4 \\

\rowcolor{blue!15}
(ii)
& EMD
& \textbf{59.5} & \textbf{64.3}
& \textbf{59.5} & \textbf{59.9}
& \textbf{58.1} & \textbf{55.1}
& \textbf{42.1} & \textbf{53.7} \\

\bottomrule
\end{tabular}
}

\caption{
\textbf{Ablation study on the Phase 1 distribution objective in the reference-free setting.}
KL: Kullback--Leibler divergence; EMD: Earth Mover's Distance.
}
\label{tab:ablation_phase1_objective}
\end{table}

\subsection{Same-Backbone Comparison}
Table \ref{tab:same_backbone_comparison} presents the quantitative comparison results with the same-backbone baselines. We implemented G-VEval and FLEUR using the same Qwen3-VL-2B backbone as \textsc{Rigel}. Table \ref{tab:same_backbone_comparison} shows that original \textsc{Rigel} outperformed the same-backbone G-VEval and FLEUR baselines across all benchmarks. These results indicate that the performance gains do not stem from the backbone or prompt engineering, but instead primarily arise from the proposed self-distillation pipeline.

\subsection{Same-Data Comparison}
To disentangle the contribution of the training data from that of the proposed method, we conducted controlled evaluations for both video and image captioning.
Table \ref{tab:same_data_baselines} presents the quantitative results with same-data baselines in image captioning.
As shown in Table \ref{tab:same_data_baselines}, \textsc{Rigel} outperformed both baselines on all benchmarks. Specifically, \textsc{Rigel} outperformed Pearl by 5.7, 0.4, and 1.8 points on Composite, Flickr8K-Expert, and Flickr8K-CF, respectively, and outperformed DENEB trained on Spica by 7.5, 2.1, and 2.2 points on the same benchmarks, respectively.

Table \ref{tab:same_data_baselines_vatex} shows the comparison with same-data baselines in video captioning.
We fine-tuned PAC-S++ and EMScore on Vid-Lepus because no supervised baseline currently exists.
As shown in Table \ref{tab:same_data_baselines_vatex}, our proposed metric achieved 42.1 and 53.7 for $\tau_b$ and $\rho$, respectively, outperforming the fine-tuned PAC-S++ by 10.9 and 13.4 points and the fine-tuned EMScore by 11.9 and 14.7 points on VATEX-EVAL.
These results indicate that the performance gains of \textsc{Rigel} are not attributable solely to training data, but to the proposed method itself.
\subsection{Objective Ablation Study}

Table~\ref{tab:ablation_phase1_objective} shows the quantitative results of
the ablation study.
We investigated the effect of the Phase~1 distributional objective by comparing
KL divergence with EMD.
EMD outperformed KL divergence across all benchmarks.
Specifically, EMD improved Composite by 0.5 points for both $\tau_b$ and
$\tau_c$, Flickr8K-EX by 0.2 points for both $\tau_b$ and $\tau_c$, and
Nebula by 0.4 points for both $\tau_b$ and $\tau_c$.
On VATEX-Eval, EMD improved $\tau_b$ and $\rho$ by 0.2 and 0.3 points,
respectively.
These results indicate that accounting for the ordinal structure of the
five-point score space provides more suitable training signals for Phase~1
distillation.

\section{Error Analysis}
\label{sec:error_analysis}
To investigate the limitations of the proposed metric, we analyzed cases where the proposed metric failed to perform as expected.
We defined failure cases as samples where the absolute difference between $y$ and $\hat{y}$ exceeded $0.5$, where y denotes the normalized human judgment score. We identified $45$ and $182$ failure cases in the test set of the Nebula and VATEX-EVAL dataset, respectively. 

Table ~\ref{tab:failure_modes_nebula} shows the results of the error analysis.
We identified 6 major failure modes:
\begin{itemize}[leftmargin=10pt,itemsep=0pt]
    \item \underline{Overestimated underspecified captions}: This category includes cases where the metric gave scores that were higher than human judgment scores to candidates that were correct but lacked sufficient detail.
    \item \underline{Overestimated inaccurate captions}: This category refers to cases where the metric gave high scores to candidates containing incorrect descriptions of objects, actions, or scenes.
    \item \underline{Overpenalized alternative focuses}: This category includes cases where the candidate focused on visual content that differed from the references and the metric gave scores lower than the human judgments.
    \item \underline{Overestimated alternative focuses}: This category refers to cases where the candidate focused on visual content that differed from the references and missed the main content, but the metric gave a score higher than the human judgments.
    \item \underline{Overpenalized local errors}: This category includes cases where the metric excessively penalized candidates for minor errors, even though the overall caption remained acceptable.
    \item \underline{Annotation error}:  This category includes samples where the human judgments were inappropriate.
\end{itemize}

\begin{table}[t]
  \centering
  \begin{tabular}{lc}
    \toprule
    Description & Count \\
    \midrule
    Overestimated underspecified captions & 14 \\
    Overestimated inaccurate captions & 8 \\
    Overpenalized alternative focuses  & 2 \\
    Overestimated alternative focuses & 6 \\
    Overpenalized local errors & 2 \\
    Annotation error & 13 \\
    \midrule
    Total & 45 \\
    \bottomrule
  \end{tabular}
  \caption{Categorization of the failure modes on Nebula. We analyzed the 45 samples with the greatest absolute differences between $\hat{y}$ and $y$.}
  \label{tab:failure_modes_nebula}
\end{table}

\begin{table}[t]
  \centering
  \begin{tabular}{lc}
    \toprule
    Description & Count \\
    \midrule
    Underestimated accurate captions & 29 \\
    Underestimated lexical variants & 43 \\
    Underestimated alternative focuses & 16 \\
    Overestimated inaccurate captions & 3 \\
    Overpenalized local errors & 4 \\
    Annotation error & 5 \\
    \midrule
    Total & 100 \\
    \bottomrule
  \end{tabular}
  \caption{Categorization of the failure modes on VATEX-EVAL. We analyzed the 100 samples with the greatest absolute differences between $\hat{y}$ and $y$.}
  \label{tab:failure_modes}
\end{table}
Table ~\ref{tab:failure_modes} shows the results of the error analysis on VATEX-EVAL. Among the 182 VATEX-EVAL failure cases, we manually analyzed the 100 cases with the largest absolute errors.
We identified 6 major failure modes:

\begin{itemize}[leftmargin=10pt,itemsep=0pt]
    \item \underline{Underestimated accurate captions}: This category encompasses cases where the proposed metric gave scores lower than those of human judgment to candidates that accurately describe the video.
    \item \underline{Underestimated lexical variants}: This category refers to cases where the proposed metric underestimated candidates that used expressions different from the references, even though they preserved the meaning.
    \item \underline{Underestimated alternative focuses}: This category includes cases where the proposed metric gave lower scores to candidates that focused on valid scenes, objects, or actions different from those emphasized in the references.
    \item \underline{Overestimated inaccurate captions}: This category refers to cases where the proposed metric gave scores higher than the human judgments to captions containing incorrect descriptions.
    \item \underline{Overpenalized local errors}: This category includes cases where the proposed metric excessively penalized captions for minor errors, even when the overall caption remained largely appropriate.
    \item \underline{Annotation error}: This category includes samples where the human judgments were inappropriate.
\end{itemize}

Table~\ref{tab:failure_modes} shows that the primary cause of errors was underestimation, mainly caused by the limited coverage of the reference captions.
These errors likely arise because the metric places strong emphasis on references and does not fully exploit visual evidence in this setting.
Consequently, visually consistent candidates can be underestimated when they differ from the references in wording or focus.
In future work, we plan to extend the metric by introducing a visual-token compression mechanism that enables the model to incorporate richer visual evidence~\cite{auroracap,timesuite}.

\section{Prompts in \textsc{Rigel}}
\label{sec:prompts}

This section provides the full prompts for all four settings: reference-free and reference-based video captioning, and reference-free and reference-based image captioning.

\TColorBox{Full prompt for video captioning in the reference-free setting}{
Evaluate the quality of a video caption based on video frames.\\
\vspace*{1\baselineskip}
Evaluation Criteria:\\
- Score ranges from 1 to 5\\
- 1: Completely incorrect or irrelevant\\
- 5: Perfectly describes the video content\\
- Penalize redundancies and irrelevant information\\
- Score 1 if the generated caption contains any hallucination\\
\vspace*{1\baselineskip}
Evaluation Steps:\\
1. Examine the video frames to understand the main content\\
2. Assess how accurately the caption describes the video\\
3. Compare the generated caption to video frames\\
4. Assess coverage of main points and relevance\\
5. Assign ONE score from 1 to 5\\
\vspace*{1\baselineskip}
Generated Caption:\\
{cand}\\
\vspace*{1\baselineskip}
Please output only a single integer from 1 to 5, without any explanation or formatting.\\
Your score is
}
\TColorBox{Full prompt for video captioning in the reference-based setting}{
Evaluate the quality of a video caption based on video frames and reference captions.\\
\vspace*{1\baselineskip}
Evaluation Criteria:\\
- Score ranges from 1 to 5\\
- 1: Completely incorrect or irrelevant\\
- 5: Perfectly describes the video content with all key information\\
- Penalize redundancies and irrelevant information\\
- Score 1 if the generated caption contains any hallucination\\
\vspace*{1\baselineskip}
Evaluation Steps:\\
1. Examine the video frames to understand the main content\\
2. Assess how accurately the caption describes the video\\
3. Compare the generated caption to both video frames and references\\
4. Assess coverage of main points and relevance\\
5. Assign ONE score from 1 to 5\\
\vspace*{1\baselineskip}
Reference Captions:\\
{refs\_text}\\
\\
Generated Caption:\\
{cand}\\
\\
Please output only a single integer from 1 to 5, without any explanation or formatting.\\
Your score is\\

}
\TColorBox{Full prompt for image captioning in the reference-free setting}{
Evaluate the quality of a image caption based on image.\\
\vspace*{1\baselineskip}
Evaluation Criteria:\\
- Score ranges from 1 to 5\\
- 1: Completely incorrect or irrelevant\\
- 5: Perfectly describes the image content\\
- Penalize redundancies and irrelevant information\\
- Score 1 if the generated caption contains any hallucination\\
\\
Evaluation Steps:\\
1. Examine the image to understand the main content\\
2. Assess how accurately the caption describes the image\\
3. Compare the generated caption to image\\
4. Assess coverage of main points and relevance\\
5. Assign ONE score from 1 to 5\\
\\
Generated Caption:\\
{cand}\\
\\
Please output only a single integer from 1 to 5, without any explanation or formatting.\\
Your score is\\

}
\TColorBox{Full prompt for image captioning in the reference-based setting}{
Evaluate the quality of a image caption based on image and reference captions.\\
\vspace*{1\baselineskip}
Evaluation Criteria:\\
- Score ranges from 1 to 5\\
- 1: Completely incorrect or irrelevant\\
- 5: Perfectly describes the image content with all key information\\
- Penalize redundancies and irrelevant information\\
- Score 1 if the generated caption contains any hallucination\\
\vspace*{1\baselineskip}
Evaluation Steps:\\
1. Examine the image to understand the main content\\
2. Assess how accurately the caption describes the image\\
3. Compare the generated caption to both image and references\\
4. Assess coverage of main points and relevance\\
5. Assign ONE score from 1 to 5\\
\vspace*{1\baselineskip}
Reference Captions:\\
{refs\_text}\\
\vspace*{1\baselineskip}
Generated Caption:\\
{cand}\\
\vspace*{1\baselineskip}
Please output only a single integer from 1 to 5, without any explanation or formatting.\\
Your score is\\

}

\section{Additional Details for ARR Checklist}
\paragraph{Discuss the License for Artifacts.}
\textsc{Rigel} and Vid-Lepus are released under the BSD 3-Clause Clear License.
The licenses of the models and datasets used in
this study are summarized below:

\begin{flushleft}
\textbf{Qwen3-VL}~\cite{qwen3vl}:

\hfill Apache 2.0 license

\vspace{0.5em}

\textbf{Qwen2.5-VL} ~\cite{qwen25vl}:

\hfill Qwen Research License Agreement

\vspace{0.5em}

\textbf{LLaVA-OneVision-1.5}~\cite{llavaov1.5}:

\hfill Apache 2.0 license

\vspace{0.5em}

\textbf{InternVL-3.5}~\cite{internvl35}:

\hfill Apache 2.0 license

\vspace{0.5em}

\textbf{Spica}~\cite{pearl}:

\hfill License not explicitly specified by the distributor

\vspace{0.5em}

\textbf{Flickr8K-Expert}~\cite{flickr}:

\hfill License not explicitly specified by the distributor

\vspace{0.5em}

\textbf{Flickr8K-CF}~\cite{flickr}:

\hfill License not explicitly specified by the distributor

\vspace{0.5em}

\textbf{Nebula}~\cite{deneb}:

\hfill License not explicitly specified by the distributor

\vspace{0.5em}

\textbf{Composite}~\cite{composite}:

\hfill License not explicitly specified by the distributor

\vspace{0.5em}

\textbf{FOIL} ~\cite{foil}:

\hfill Creative Commons Attribution 4.0 license

\vspace{0.5em}

\textbf{VATEX-EVAL}~\cite{emscore}:

\hfill MIT license

\vspace{0.5em}

\textbf{ActivityNet-Fact}~\cite{factvc}:

\hfill License not explicitly specified by the distributor

\vspace{0.5em}

\textbf{YouCook2-Fact}~\cite{factvc}:

\hfill License not explicitly specified by the distributor

\vspace{0.5em}

\textbf{ActivityNet-FOIL}~\cite{emscore}:

\hfill MIT license
\end{flushleft}
\paragraph{Artifact Use Consistent With Intended Use.}
All existing artifacts used in this study were employed in accordance with their intended use. For the artifacts developed in this study, we define their intended use as general academic and research purposes, consistent with the original access conditions of the datasets and models used.
\paragraph{Data Contains Personally Identifying Info Or Offensive Content.}
The collected data do not contain personally identifiable or offensive content. All data used in this study are publicly available. We further confirmed that the source websites, repositories, and publications contain no statements raising concerns about personal information.
\end{document}